\documentclass[10pt,twocolumn,letterpaper]{article}

\usepackage{cvpr}
\usepackage{times}
\usepackage{epsfig}
\usepackage{graphicx}
\usepackage{amsmath}
\usepackage{amssymb}

\usepackage{multirow}
\usepackage{subfigure}
\usepackage{color}
\usepackage{algorithm}
\usepackage[noend]{algpseudocode}
\usepackage[breaklinks=true,bookmarks=false]{hyperref}

\cvprfinalcopy 

\def\cvprPaperID{2532} 

\ifcvprfinal\fi
\begin{document}

\title{Few-shot Adaptive Faster R-CNN}

\author{\normalsize{Tao~Wang$^1$} \qquad \quad  \normalsize{Xiaopeng~Zhang$^{1,2}$}  \quad  \qquad \normalsize{Li~Yuan$^1$}  \quad  \qquad \normalsize{Jiashi~Feng$^1$}\\
	\small{$^{1}$Department of Electrical and Computer Engineering, National University of Singapore, Singapore} \\
	\small{$^{2}$Huawei Noah's Ark Lab, Shanghai, China} \\
	{\small \tt twangnh@gmail.com}  \ \  {\small \tt zhangxiaopeng12@huawei.com}  \ \ {\small\tt ylustcnus@gmail.com} \ \ {\small\tt elefjia@nus.edu.sg}
}

\maketitle
\thispagestyle{empty}

\begin{abstract}
To mitigate the detection performance drop caused by domain shift, we aim to develop a novel few-shot adaptation approach that requires only a few target domain images with limited bounding box annotations. To this end, we first observe several significant challenges. First, the target domain data is highly insufficient, making most existing domain adaptation methods ineffective. Second, object detection involves simultaneous localization and classification, further complicating the model adaptation process. Third, the model suffers from over-adaptation (similar to overfitting when training with a few data example) and instability risk that may lead to degraded detection performance in the target domain. To address these challenges, we first introduce a pairing mechanism over source and target features to alleviate the issue of insufficient target domain samples. We then propose a bi-level module to adapt the source trained detector to the target domain: 1) the split pooling based image level adaptation module uniformly extracts and aligns paired local patch features over locations, with different scale and aspect ratio; 2) the instance level adaptation module semantically aligns paired object features while avoids inter-class confusion. Meanwhile, a source model feature regularization (SMFR) is applied to stabilize the adaptation process of the two modules. Combining these contributions gives a novel few-shot adaptive Faster-RCNN framework, termed FAFRCNN, which effectively adapts to target domain with a few labeled samples. Experiments with multiple datasets show that our model achieves new state-of-the-art performance under both the interested few-shot domain adaptation(FDA) and unsupervised domain adaptation(UDA) setting.
\end{abstract}

\begin{figure}[t]
	\begin{center}
	\includegraphics[width=1\linewidth]{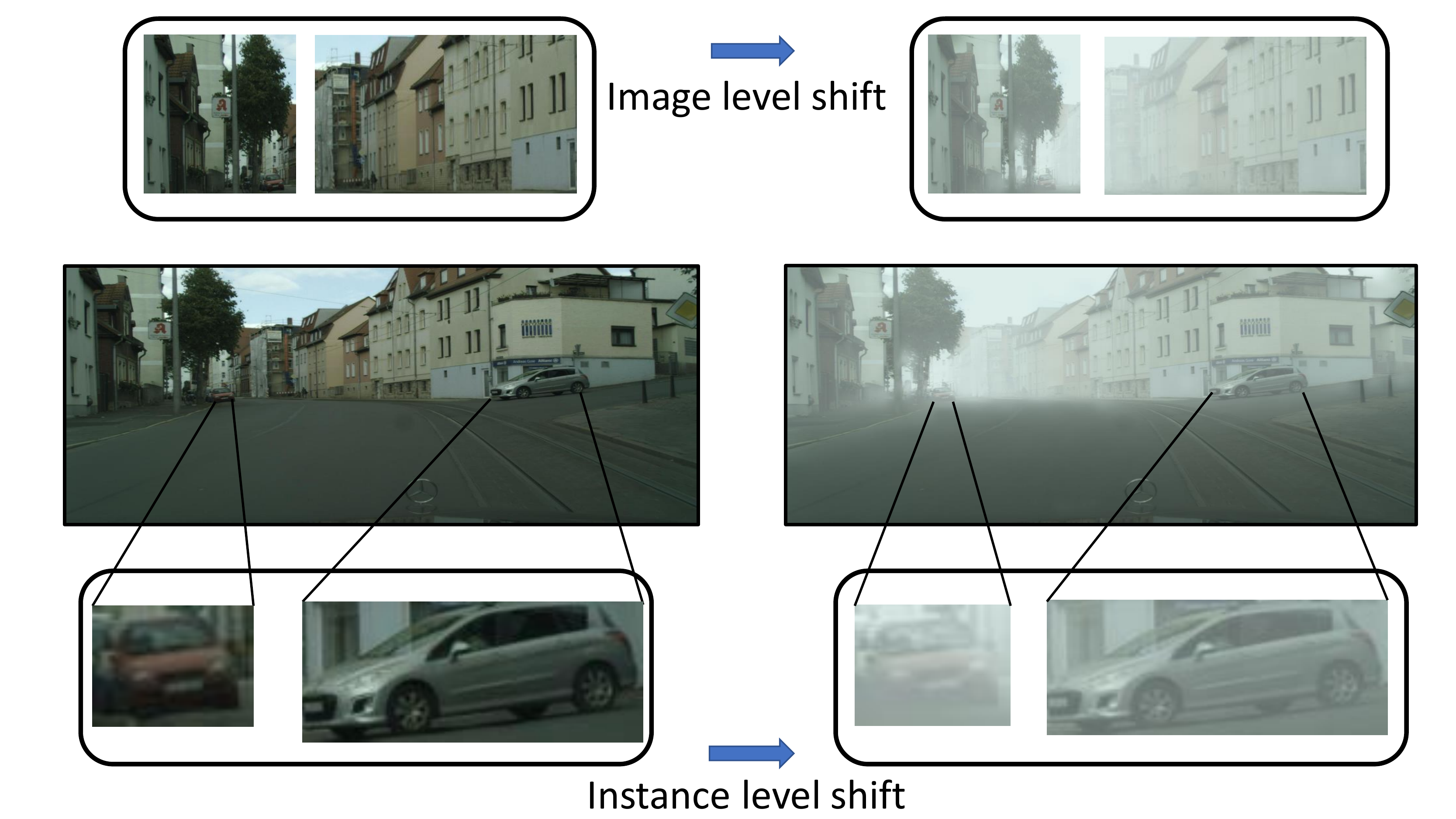}
	\end{center}
	\caption{Illustration of our main idea. Middle row are two images from Cityscapes and Foggy Cityscapes respectively. Top row shows background patches reflecting image level domain shift and bottom row shows independent objects (cars) reflecting object instance level domain shift. 
	}  
	\label{idea}
	\vspace{-15pt}
\end{figure}

\vspace{-10pt}
\section{Introduction}
Humans can easily recognize familiar objects from new domains, while current object detection models suffer significant performance drop in unseen environments due to domain shift. Poor adaptability to new domains severely limits the applicability and efficacy of these models. 
Previous works tackling domain shift issues for deep CNN models~\cite{ganin2016domain, tzeng2015simultaneous, long2015learning, bousmalis2017unsupervised} are mainly targeted at the unsupervised domain adaptation (UDA) setting, which requires a large amount of target domain data and  comparatively long adaptation time. Only a few works consider the supervised domain adaptation (SDA)~\cite{saha2011active, conjeti2016supervised, motiian2017unified} setting. However, as UDA methods, they mainly focus on the simple task of classification, and may not apply well to more complex tasks like object detection that involves localizing and classifying all individual objects over high resolution inputs. 

In this paper, we explore the possibility of adapting an object detector trained with source domain data to target domain \emph{with only a few loosely annotated target image samples} (not all object instances are annotated). This is based on our key observation that limited target samples can still largely reflect major domain characteristics, \textit{e.g.} illumination, weather condition, individual object appearance, as shown in Fig.~\ref{idea}. Also, the setting is appealing in practice as collecting a few representative data from a new domain needs negligible effort, meanwhile can reduce
the inevitable noise brought by large amount of samples. However, it is very challenging to learn domain invariant representation with only a few target data samples, and detectors require fine-grained high resolution features for reliable localization and classification.  

To address this challenge, we proposed a novel framework that consists of two level of adaptation modules coupled with a feature pairing mechanism and a strong regularization for stable adaptation. The pairing process pairs feature samples into two groups to effectively augment limited target domain data, pairs in first group consist of one sample from target domain and  one from the source domain, and pairs in the second group are both from the source domain. Similar approach has been used in ~\cite{motiian2017few} for augmenting image samples, while we augment local feature patches and object features in the two adaptation module respectively. With the introduced pairing mechanism,
the image-level module uniformly extracts and aligns paired multi-grained patch features to address the global domain-shift like illumination; the instance-level module semantically matches paired object features while avoids confusion between classes as well as reduced discrimination ability. Both of these two modules are trained with a domain-adversarial learning method. 
We further propose a strong regularization method, termed source model feature regularization (SMFR), to stabilize training and avoid over-adaptation by imposing consistency between source and adapted models on feature response of foreground anchor locations. The bi-level adaptation modules combined with SMFR  can robustly adapt source trained detection model to new target domain with only few target sample data. The resulted framework, termed few shot adaptive Faster R-CNN (FAFRCNN), offers a number of advantages:
\begin{itemize}
	\setlength\itemsep{0em}
\item \textbf{Fast adaptation.} For a source trained model, our framework empirically only needs hundreds step of adaptation updates to reach desirable performance under all established scenarios. In contrast previous methods under UDA setting~\cite{tzeng2017adversarial, chen2018domain} requires tens of thousands of steps to train.

\item \textbf{Less data collection cost.} With only few representative data sample, the FAFRCNN model can greatly boost source detector on target domain, drastically mitigating data collection cost. Under the devised loosely annotation process, the amount of human annotating time is reduced significantly.

\item \textbf{Training stability.} Fine-tuning with limited target data sample can lead to severe over-fitting. Also, domain adaptation approaches relying on adversarial objective might be unstable and sensitive to   initialization of model parameters. This issue  greatly limits their applicability. The proposed  SMFR approach  enables  the model to avoid over-fitting and   benefit from the few target data samples. For the two adversarial adaptation modules, although imposing SMFR could not significantly boost their performance, the variance over different runs is drastically reduced. Thus SMFR provides much more stable and reliable model adaptation.
\end{itemize}

To demonstrate the efficacy of the proposed FAFRCNN for cross-domain object detection, we conduct the few-shot adaptation experiments under various scenarios constructed with multiple datasets including Cityscapes, SIM10K, Udacity self-driving and Foggy Cityscapes. Our model significantly surpasses compared methods and outperforms state-of-art method using full target domain data. When applied to UDA setting, our method generates new state-of-art result for various scenarios.

\section{Related Work}

\paragraph{Object Detection} Recent years have witnessed remarkable progress on object detection with deep CNNs  and various large-scale datasets. Previous detection architectures are grouped into two- or multi-stage models like R-CNN~\cite{girshick2014rich}, Fast R-CNN~\cite{girshick2015fast}, Faster R-CNN~\cite{ren2015faster} and Cascaded R-CNN~\cite{cai2017cascade}, as well as single-stage models like YOLO~\cite{redmon2016you}, YOLOv2~\cite{redmon2017yolo9000}, SSD~\cite{liu2016ssd} and Retinanet~\cite{lin2018focal}. However, all of them require a large amount of training data with careful annotations, thus are not directly applicable to object detection in unseen domains.


\vspace{-4mm}
\paragraph{Cross-domain Object Detection}
Recent works on domain adaptation with CNNs mainly address the simple task of classification~\cite{long2015learning, ganin2014unsupervised, ghifary2016deep, busto2017open, li2017deeper, haeusser2017associative, lu2017unsupervised}, and only a few consider object detection.~\cite{xu2014domain} proposed a framework to mitigate the domain shift problem of deformable part-based model (DPM).~\cite{raj2015subspace} developed subspace alignment based domain adaptation for the R-CNN model. A recent work  \cite{inoue2018cross} used a two-stage iterative domain transfer and pseudo-labeling approach to tackle cross-domain weakly supervised object detection.~\cite{chen2018domain} designed three modules for unsupervised domain adaptation of the object detector. In this work, we aim at adapting object detectors with a few target image samples and build a framework for robust adaptation of state-of-the-art Faster R-CNN models under this setting.
\vspace{-7mm}
\paragraph{Few-shot Learning}
Few-shot learning~\cite{fei2006one} was proposed to learn a new category with only a few examples, just as humans do. Many works are based on Bayesian inference~\cite{lake2013one, lake2015human}, and some leverage memory machines~\cite{graves2014neural, santoro2016one}. Later,~\cite{hariharan2017low} proposed to transfer the base class feature to a new class; a recent work~\cite{finn2017model} proposed a meta learning based approach which achieves state-of-the-art. Incorporating few-shot learning into object detection was previously explored.~\cite{dong2017few} proposed to learn an object detector with a large pool of unlabeled images and only a few annotated images per category, termed few-shot object detection (FSOD);~\cite{chen2018lstd} tackled the setting of few-shot object detection with a low-shot transfer detector (LSTD) coupled with designed regularization. Our FDA setting differs in that target data distribution changed but task remain the same, while few-shot learning aims at a new tasks.

\begin{figure*}[!t]
	\centering
	\includegraphics[width=0.95\linewidth, height = 0.35\linewidth]{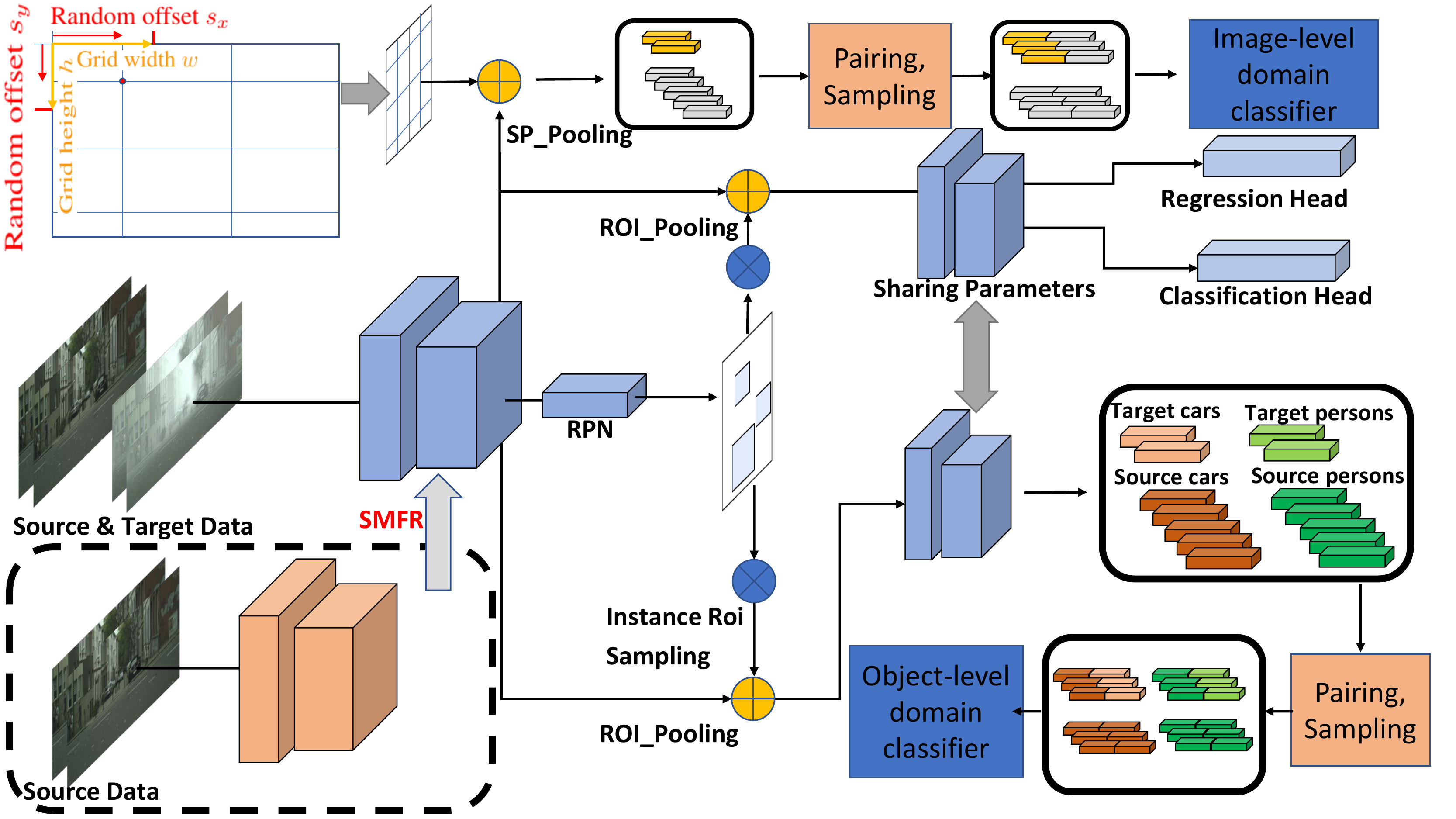}
	\caption{Framework of the proposed few shot adversarial adaptive Faster R-CNN model(FAFRCNN). We address the domain shift with image level and instance level adaptation modules, the former with different grid size adapts multi-grained feature patches and latter semantically aligns independent object appearance, the modules augmented with the proposed pairing mechanism result in effective alignment of feature representation in such few shot scenario(refer to Section~\ref{method_section} for details), we further developed source model feature regularization(SMFR) which dramatically stabilizes the adaptation process.}
	\label{framework}
	\vspace{-10pt}	
\end{figure*}

\section{Method}\label{method_section}
\vspace{-2pt}
In this section,  we elaborate on our proposed few-shot domain adaptation approach for detection.  To tackle the issue brought by insufficient target domain samples, we introduce a novel feature pairing mechanism built upon features sampled by split pooling and instance ROI sampling. Our proposed approach performs domain adaptation over the paired features at both image and object-instance levels through domain-adversarial learning, where the first level alleviates global domain shift and the second level semantically aligns object appearance shift while avoiding confusion between classes. To stabilize the training and avoid over-adaptation, we finally introduce the source model feature regurgitation technique. We apply these three novel techniques to Faster R-CNN model and obtain the few-shot adaptive Faster-RCNN (FAFRCNN),  which is able to adapt to novel domains with only a few target domain examples. 
\vspace{-0.5pt}
\subsection{Problem Setup}

Suppose we have a large set of source domain training data $ (X_S, Y_S)$ and a very small set of target data $ (X_T, Y_T)$, where $X_S$ and $X_T$ are input images, $Y_S$ denotes complete bounding box annotation for $X_S$, and $Y_T$ denotes loose annotation for $X_T$.  With only a few object instances in the target domain images annotated, our goal is to adapt a detection model trained on source training data to the target domain with minimal performance drop. We only consider loose bounding box annotation to reduce annotation effort.




\subsection{Image-level Adaptation}

%

Inspired by the superior result of the patch based domain classifier compared to its full image counterpart in previous seminal works~\cite{isola2017image, zhu2017unpaired} for image to image translation. We propose \textbf{split pooling} (SP) to uniformly extract local feature patches across locations with different aspect ratio and scale for domain adversarial alignment.


Specifically, given grid width $ w $ and height $ h $, the proposed split pooling  first generates random offsets $s_x$ and $s_y$ for $x$- and $y$-axis ranging from $0$ to the full grid width $w$ and height $h$ respectively (\textit{i.e.}, $0{<}s_x{<}w, 0{<}s_y{<}h, s_x,s_y\in \mathbb{N}$), as shown in the top left panel of Fig.~\ref{framework}. A random grid is formed on the input image with the offset of $(s_x,s_y)$ starting from the top left corner of the input image. This random sampling scheme gives a trade-off between static grid that may generate biased sampling, and exhausting all grid locations that suffers redundancy and over-sampling.

The grid window width $ w $ and height $ h $ are set with scales and ratios as anchor boxes in Faster R-CNN. We empirically choose 3 scales (large scale 256, medium scale 160, and small scale 96, corresponding to feature size 16, 10 and 6 on relu\_5\_3 of VGG16 network) and 3 aspect ratios (0.5, 1, 2), resulting in 9 pairs of $ w $ and $ h $. For each pair, gird is generated then non-border rectangles in the grid are pooled into fixed sized features with ROI pooling.
Pooling enables different sized grids to be compatible with a single domain classifier without changing the patch-wise characteristics of the extracted features. 
Formally, let $ f $ be the feature extractor  and $ X $ be the set of input images. We perform split pooling at three scales, result in the features $ sp_{l}(f(X))$, $sp_{m}(f(X))$, and $sp_{s}(f(X)) $ respectively. We separate them according to scales as we want to investigate the contribution of different scales independently.
These local patch features can reflect image-level domain shifts like varied illumination, weather change, \textit{etc}. Since those shifts spread on the whole image, the phenomenon is more evident for object detection as input images are usually large. 

We then develop image-level adaptation module which performs \textbf{multi-scale alignment with paired local features}. 
Specifically, it tackles image-level shift by first pairing the extracted local features from split pooling to form two groups for each of the three scales. \emph{e.g.}, for the small scale patch, $ G_{s_1} = \{(g_s, g_s)\}, G_{s_2} = \{(g_s, g_t)\} $, where $ g_s\sim sp_{s}(f(X_S)) $ and $ g_t\sim sp_{s}(f(X_T)) $. Here the pairs within the first group $G_{s_1}$ consist of samples from the source domain only, and pairs within the second group $G_{s_2}$ consist of one sample from source and another from the target  domain. Such a pairing scheme effectively augments the limited target domain feature samples. 

To adapt the detection model, domain-adversarial learning objective  is imposed to align the constructed two groups of features. The domain-adversarial learning~\cite{ganin2014unsupervised, tzeng2015simultaneous, tzeng2017adversarial} employs the principle in generative adversarial learning~\cite{goodfellow2014generative} to minimize an approximated domain discrepancy distance through adversarial objective on feature generator and domain discriminator. Thus the data distribution is aligned and source task network can be employed for the target domain. Specifically the domain discriminator tries to classify the feature to source and target domain while the feature generator tries to confuse the discriminator.
The learning objective of small scale discriminator $D^{sp_{s}}$ is to minimize
\begin{equation*}
\begin{aligned}
L_{sp{_{sd}}} = -&\mathbb{E}_{x\sim G_{s_1}}[\log D^{sp_{s}}(x)] \\ 
&-\mathbb{E}_{x\sim G_{s_2}}[\log (1-D^{sp_{s}}(x))],
\end{aligned}
\label{im_ds}
\end{equation*}
such that the discriminator can tell clearly the source-source feature pairs apart from source-target feature pairs. 
The objective of the generator  is to transform the features from both domains such that they are not distinguishable to the discriminator, by maximizing the above loss. 


We can similarly get losses for medium and large scale discriminator as $ L_{sp{_{md}}} $ and $ L_{sp{_{ld}}} $. We use 3 separate discriminators for each scale. In addition, this module operates requiring no supervision. Thus it can be used for unsupervised domain adaptation (UDA). Together, the image level discriminator's objective is to minimize:
\begin{equation*}
\begin{aligned}
L_{im_d} = L_{sp{_{sd}}} + L_{sp{_{md}}}+ L_{sp{_{ld}}},
\end{aligned}
\label{im_d}
\end{equation*}
and the feature generator's objective is to maximize $ L_{im_d} $.

\subsection{Instance-level Adaptation}
To mitigate object instance level domain shift, we propose the instance-level adaptation module which 
\textbf{semantically aligns paired object features}. 

Specifically, we extend the Faster R-CNN ROI sampling to instance ROI sampling. The Faster R-CNN ROI sampling scheme samples ROIs to create training data for classification and regression heads. It by default separates foreground and background ROIs with an IOU threshold of 0.5 and samples them at a specific ratio (\emph{e.g.}, 1:3). Differently,  our proposed instance ROI sampling keeps all the foreground ROIs with higher IOU threshold (\emph{i.e.}, 0.7 in our implementation) to ensure the ROIs are closer to real object regions and suitable for alignment. The foreground ROI features of source and target domain images, according to their class, are passed through the intermediate layers (\emph{i.e.}, the layers after ROI pooling but before classification and regression heads) to get sets of source object features $ O_{is} $ and target object features $ O_{it} $. Here  $ i\in[0,C] $ is the class label and $ C $ is the total number of classes. Then they are further paired into two groups the same way as image level patch features, resulting in $ N_{i1}=\{(n_{is}, n_{is})\}$ and $ N_{i2}=\{(n_{is}, n_{it})\} $. Here $ n_{is}\sim O_{is} $ and $ n_{it}\sim O_{it} $. The multi-way instance-level discriminator $D^{ins}$ has $ 2\times C $ outputs with a following objective to minimize:
\begin{equation*}
\begin{aligned}
L_{ins_d} = \sum_{i=1}^C - & \mathbb{E}_{x\sim N_{i1}}[\log D^{ins}(x)_{i1}] \\ 
& - \mathbb{E}_{y\sim N_{i2}}[\log D^{ins}(y)_{i2}].
\end{aligned}
\label{ins_g}
\end{equation*}
Here $D^{ins}(x)_{i1}$ denotes discriminator output over the $i$-th class of first group. Correspondingly, the objective of feature generator is to minimize
\begin{equation*}
\begin{aligned}
L_{ins_g} = \sum_{i=1}^C - &\mathbb{E}_{x\sim N_{i1}}[\log D^{ins}(x)_{i2}] \\ 
& - \mathbb{E}_{y\sim N_{i2}}[\log D^{ins}(y)_{i1}],
\end{aligned}
\label{ins_dc}
\end{equation*}
which aims to confuse the discriminator between two domains while avoid misclassification to other classes. 

\subsection{Source Model Feature Regularization} \label{SMFR_method} 
Training instability is a common issue for adversarial learning and is more severe for cases of insufficient training data, which may result in over-adaptation. Fine-tuning with limited target data would also unavoidably lead to overfitting. We resort to a strong regularization to address the instability by forcing the adapted model to produce consistent feature response on source input with the source model in the sense of $\ell_2$ difference.  The purpose is to avoid over-updating learned representation towards limited target samples that degrades the performance. A similar form of $\ell_2$ penalty on the feature map was used in image to image translation method~\cite{bousmalis2017unsupervised, isola2017image} to constrain content change.

Formally, Let $f_{s}$ and $f_{t}$ be the feature extractors of the source model and the adapted model respectively. Then the source model feature regularization (SMFR)  term is
\begin{equation*}
\begin{aligned}
L_{reg}=\mathbb{E}_{x_{s}\sim X_{S}}\frac{1}{wh}||f_{s}(x_{s})-f_{t}(x_{s})||_{2}^{2},
\end{aligned}
\label{loss_reg}
\end{equation*}
where $w$ and $h$ are the width and height of the feature map.

However, object detection cares more about local foreground feature regions while background area is usually unfavorably dominant and noisy. We find directly imposing the regularization on global feature map leads to severe deterioration when adapting to the target domain. Thus we propose to estimate those foreground regions on the feature map as the anchor locations that have IOU with ground truth boxes larger than a threshold (0.5 is used in implementation). Denote $M$ as the estimated foreground mask. Then we modify the proposed regularization as follows:
\begin{equation*}
\begin{aligned}
L_{reg}=E_{x_{s}\sim X_{S}}\frac{1}{k}||(f_{s}(x_{s})-f_{t}(x_{s}))*M||_{2}^{2},
\end{aligned}
\label{smfr}
\end{equation*}
where $k$ is the number of positive mask locations. This is partially inspired by the ``content-similarity loss'' from~\cite{bousmalis2017unsupervised} that employs available rendering information to impose $\ell_2$ penalty on foreground regions of the generated image.

\subsection{Training of FAFRCNN} 
The framework is initialized with the source model and optimized by alternating between following objectives:
\textbf{Step 1}. Minimize the following loss w.r.t. full detection model:
	$L_g = \alpha(L_{im_g}+L_{ins_g})+\beta L_{det}+\lambda L_{reg}$,
	where $ L_{det} $ denotes Faster R-CNN detection training loss on source data, $\alpha$, $ \beta $ and $ \lambda $ are balancing hyperparameters controlling interaction between losses.
	\textbf{Step 2}.  Minimize following loss w.r.t. domain discriminators:
	$L_d = L_{im_d}+L_{ins_d}$.
%
%

\vspace{-1mm}
\section{Experiments}\label{experiments}
In this section, we present evaluation results of the proposed method on adaptation scenarios capturing different domain shift constructed with multiple datasets.
In experiments, VGG16 network based Faster-RCNN is used as the detection model. 
\vspace{-1mm}
\subsection{Datasets and Setting} \label{sec_data}

\paragraph{Datasets} We adopt following four datasets to establish the cross-domain adaptation  scenarios for evaluating the adaptation ability of our model and comparing methods. 
	The \textbf{SIM10K}~\cite{johnson2016driving} dataset contains 10k synthetic images with bounding box annotation for car, motorbike and person.
	The \textbf{Cityscapes}  dataset contains around 5000 accurately annotated real world images with pixel-level category labels. Following~\cite{chen2018domain}, we take box envelope of instance mask for bounding box annotations. The \textbf{Foggy Cityscapes}~\cite{SDV18} dataset is generated from Cityscapes with simulated fog.
	The \textbf{Udacity self-driving} dataset (Udacity for short)~\cite{udacity} is an open source dataset collected with different illumination, camera condition and surroundings as Cityscapes.
\vspace{-4mm}
\paragraph{Evaluation scenarios} The established cross-domain adaptation scenarios include \textbf{Scenario-1}: SIM10K to Udacity (S $ \rightarrow $ U); \textbf{Scenario-2}: SIM10K to Cityscapes (S {$\rightarrow$} C); \textbf{Scenario-3}: Cityscapes to  Udacity (C $ \rightarrow $ U); \textbf{Scenario-4}: Udacity to Cityscapes (U $ \rightarrow $ C); \textbf{Scenario-5}: Cityscapes to Foggy Cityscapes (C $ \rightarrow $ F). 
The first two scenarios capture synthetic to real data domain shift, which is important as learning from synthetic data is very promising way to address the lack of labeled training data~\cite{christiano2016transfer, rusu2016sim, qiu2016unrealcv}; Scenario-3 and Scenario-4  constructed with both real world collected datasets mainly aim for domain shift like illumination, camera condition, etc., which is important for practical applications; And the last scenario captures the extreme weather change of normal to foggy condition. We sample from target train set and test on target val set, the source model is trained with full source dataset.

\vspace{-4.5mm}

\paragraph{Baselines} We compare our method with following baselines: (1) Source training model. The model trained with source data only and directly evaluated on target domain data. (2) ADDA~\cite{tzeng2017adversarial}. ADDA is a general framework for addressing unsupervised adversarial domain adaptation. Last feature map is aligned in experiments.
(3) Domain transfer and Fine-tuning (DT+FT). The method has been used as a module in ~\cite{inoue2018cross} for adapting object detector to target domain. In UDA setting, we use CycleGAN~\cite{zhu2017unpaired} to train and transform source image to target domain. In FDA setting, since very few target domain samples are available, we employ method in~\cite{johnson2016perceptual} that needs only one target style  image to train the transformation. This baseline is  denoted as DT$ _{f} $+FT.
(4) Domain Adaptive Faster R-CNN~\cite{chen2018domain}. The method is deliberately developed for unsupervised domain adaptation, denoted as FRCNN\_UDA.
\vspace{-0.5mm}
\subsection{Quantitative Results} \label{Quantitative_r}
We evaluate the proposed method by conducting extensive experiments on the established scenarios.
To quantify the relative effect of each step, the performances of are examined with different configurations. We also evaluate proposed split pooling based image level adaptation in the unsupervised domain adaptation (UDA) setting, where large amount of unlabeled target images are available. 

Specifically, for the few-shot domain adaptation (FDA) setting, we perform the following steps for each run: (1) Randomly sample fixed number of target domain images, ensure that needed class are presented; (2) Simulate \textit{loosely annotating process} to get annotated target domain images, i.e., only randomly annotate fixed number of object instances; (3) Gradually combine each component of our method, run the adaptation and record performance (AP); (4) Run compared methods on the same sampled images and record performance. For the UDA setting,  only proposed split pooling based adaptation component is used as no annotation is available in the target domain.




\begin{table}[]
	\small
	\renewcommand{\arraystretch}{1.3}
	\renewcommand{\tabcolsep}{2.0pt}
	\begin{tabular}{lccccccc}
		\hline
		& sp$_s$ & sp$ _m $ & sp$ _l $ & ins & ft & S$\rightarrow$U       & S $\rightarrow$C \\ \hline
		Source   &    &    &     &    &         &      34.1    & 33.5    \\ \hline
		
		\textbf{FDA setting} \\
		~~ADDA~\cite{tzeng2017adversarial}&    &    &     &    &         &  34.3$ _{\pm0.9} $  &   34.4$ _{\pm0.7} $     \\ 
		~~DT$ _{f} $+FT &    &    &     &    &         &  35.2$ _{\pm0.3} $  &   35.6$ _{\pm0.6} $     \\ 
		~~FRCNN\_UDA~\cite{chen2018domain} &    &    &    &     &    &                   33.8$ _{\pm1.0} $  &  33.1$ _{\pm0.4} $ \\ \cline{2-8} 
		\multirow{9}{*}{~~Ours} & \checkmark  &    &    &     &    &  35.1$ _{\pm0.6} $  &   35.4$ _{\pm0.8} $   \\ \cline{2-8} 
		&    & \checkmark  &    &     &    &  34.9$ _{\pm0.6} $  &   34.8$ _{\pm0.5} $   \\ \cline{2-8} 
		
		&    &    & \checkmark  &     &    &  36.0$ _{\pm1.0} $  &  34.8$ _{\pm0.8} $    \\ \cline{2-8} 
		
		& \checkmark  & \checkmark  &    &     &   &   35.2$ _{\pm1.0} $         &   35.8$ _{\pm0.6} $   \\ \cline{2-8}
		& \checkmark  & \checkmark  & \checkmark  &  &    &  36.8$ _{\pm0.6} $          &  37.0$ _{\pm0.9} $    \\ \cline{2-8} 
		&    &    &    & \checkmark   &    &  37.2$ _{\pm0.9} $          &   37.1$ _{\pm0.6} $   \\ \cline{2-8} 
		& \checkmark  & \checkmark  & \checkmark  & \checkmark   &    &  38.8$ _{\pm0.3} $          &  39.2$ _{_\pm0.5} $    \\ \cline{2-8} 
		&    &    &    &     & \checkmark  &  34.8$ _{_\pm0.4} $          & 34.6$ _{_\pm0.5} $     \\ \cline{2-8}
		& \checkmark  & \checkmark  & \checkmark  & \checkmark   & \checkmark  & \textbf{39.3$ _{_\pm0.3} $}      &  \textbf{39.8$ _{_\pm0.6} $}    \\ \hline
		\textbf{UDA setting} \\ 
		~~ADDA~\cite{tzeng2017adversarial}&    &    &     &    &         &  35.2  &   36.1     \\ 
		~~DT+FT&    &    &     &    &         & 36.1   &   36.8     \\ 
		~~FRCNN\_UDA~\cite{chen2018domain}    &    &     &   &    &         &   36.7       & 38.9  \\
		~~Ours (SP only)    &    &     &   &    &         &   \textbf{40.5}       & \textbf{41.2}  \\
		\hline
	\end{tabular}
	\vspace{2pt}
	\caption{Quantitative results of our method on Scenario-1 and Scenario-2, in terms of average precision for car detection.  UDA denotes traditional setting where large amount of unlabeled target images are available, and FDA indicates the proposed few shot domain adaptation setting. sp$ _{s} $,  sp$ _{m} $  and  sp$ _{l} $ denote  small, medium and large  scale split pooling respectively. ``ins'' indicates object instance level adaptation  and ``ft'' denotes adding fine-tuning loss with available target domain annotations. For FDA setting, both S$\rightarrow$U and S$\rightarrow$C samples 8 images per experiment round and annotate 3 car objects per image.}
	\label{SUC_table}
	\vspace{-15pt}
\end{table}

\begin{table}[]
	\small
	\renewcommand{\arraystretch}{1.3}
	\renewcommand{\tabcolsep}{2.0pt}
	\begin{tabular}{lccccccc}
		\hline
		& sp$ _s $ & sp$ _m $ & sp$ _l $ & ins & ft & C$\rightarrow $U       & U$ \rightarrow $C \\ \hline
		Source only &    &    &     &    &         & 44.5 &  44.0          \\ \hline
		\textbf{FDA setting} \\ 
		~~ADDA~\cite{tzeng2017adversarial} &    &    &     &    &         &  44.3$ _{\pm0.9} $        & 44.2$ _{\pm1.2} $   \\
		~~DT$ _{f} $+FT &    &    &     &    &         &  44.9$ _{\pm0.6} $  &   45.1$ _{\pm0.5} $     \\ 
		~~FRCNN\_UDA~\cite{chen2018domain}    &    &    &     &    &         &43.0$ _{\pm0.8} $ &            43.3$ _{\pm0.8} $  \\ \cline{2-8} 
		\multirow{9}{*}{~~Ours} & \checkmark  &    &    &     &    &45.9$ _{\pm0.7} $ &47.2$ _{\pm0.3} $     \\ \cline{2-8} 
		
		&    & \checkmark  &    &     &    &46.1$ _{\pm0.4} $ &47.6$ _{\pm0.5} $     \\ \cline{2-8} 
		
		&    &    & \checkmark  &     &    &45.3$ _{\pm0.4} $ &48.1$ _{\pm0.7} $     \\ \cline{2-8} 
		
		& \checkmark  & \checkmark  &    &     &   &  45.9$ _{\pm0.6} $& 48.0$ _{\pm0.5} $            \\ \cline{2-8}
		
		& \checkmark  & \checkmark  & \checkmark  &  &    & 46.8$ _{\pm0.3} $&48.8$ _{\pm0.9} $             \\ \cline{2-8} 
		
		&    &    &    & \checkmark   &    & 46.4$ _{\pm0.5} $&47.1$ _{\pm0.7} $             \\ \cline{2-8} 
		
		& \checkmark  & \checkmark  & \checkmark  & \checkmark   &  & 47.8$ _{\pm0.4} $ & 49.2$ _{\pm0.4} $             \\ \cline{2-8} 
		
		&    &    &    &     & \checkmark  &45.5$ _{\pm0.8} $ & 45.0$ _{_\pm0.6} $             \\ \cline{2-8}
		
		& \checkmark  & \checkmark  & \checkmark  & \checkmark   & \checkmark & \textbf{48.4$ _{\pm0.4}$}  & \textbf{50.6$_{_\pm0.6}$}             \\ \hline
		\textbf{UDA setting} \\ 
		~~ADDA~\cite{tzeng2017adversarial} &    &    &     &    &         &  46.5        & 47.5   \\
		~~DT+FT    &    &    &     &    &         &   46.1       &  47.8  \\   
		~~FRCNN\_UDA\cite{chen2018domain}    &    &    &     &    &         &    47.9      & 49.0   \\
		~~Ours (SP only)    &    &    &     &    &         &    \textbf{48.5}      & \textbf{50.2}   \\ \hline
	\end{tabular}
	\vspace{2pt}
	\caption{Quantitative results of our method on Scenario-3 and Scenario-4. For FDA setting, C$\rightarrow $U samples 16 images per experiment round, and U$\rightarrow $C samples 8 images per round, both annotate 3 car objects per image.}
	\label{CU_table}
	\vspace{-15pt}
\end{table}

\begin{table*}[]
	\small
	\centering
	\renewcommand{\tabcolsep}{1.8pt}
	\renewcommand{\arraystretch}{1.1}
	\begin{tabular}{lcccccccccccccc}
		\hline
		& sp$ _s $ & sp$ _m $ & sp$ _l $ & ins & ft & person & rider & car  & truck & bus  & train & mcycle & bicycle & mAP  \\ \hline
		Source                 &    &    &    &     &    & 24.1   & 29.9  & 32.7 & 10.9  & 13.8 & 5.0   & 14.6   & 27.9    & 19.9 \\ \hline
		\textbf{FDA setting} \\
		~~ADDA~\cite{tzeng2017adversarial}                      &    &    &    &     &    & 24.4$_{_\pm0.3}$   & 29.1$_{_\pm0.9}$  & 33.7$_{_\pm0.5}$ & 11.9$_{_\pm0.5}$  & 13.3$_{_\pm0.8}$ & 7.0$_{_\pm1.5}$   & 13.6$_{_\pm0.6}$  & 27.6$_{_\pm0.2}$    & 20.1$_{_\pm0.8}$ \\
		
		~~DT$ _{f} $+FT                     &    &    &    &     &    & 23.5$_{_\pm0.5}$   & 28.5$_{_\pm0.6}$  & 30.1$_{_\pm0.8}$ & 11.4$_{_\pm0.6}$  & 26.1$_{_\pm0.9}$ & 9.6$_{_\pm2.1}$   & 17.7$_{_\pm1.0}$   & 26.2$_{_\pm0.6}$    & 21.7$_{_\pm0.6}$ \\ 
		~~FRCNN\_UDA~\cite{chen2018domain}                &    &    &    &     &    & 24.0$_{_\pm0.8}$   & 28.8$_{_\pm0.7}$  & 27.1$_{_\pm0.7}$ & 10.3$_{_\pm0.7}$  & 24.3$_{_\pm0.8}$ & 9.6$_{_\pm2.8}$   & 14.3$_{_\pm0.8}$   & 26.3$_{_\pm0.8}$    & 20.6$_{_\pm0.8}$ \\ \cline{2-15} 
		\multirow{9}{*}{~~Ours} & \checkmark &    &    &     &    & 25.7$_{_\pm0.8}$   & 35.6$_{_\pm1.0}$  & 35.8$_{_\pm0.8}$ & 17.7$_{_\pm0.3}$  & 31.9$_{_\pm0.5}$ & 9.4$_{_\pm2.5}$   & 21.6$_{_\pm1.5}$   & 30.3$_{_\pm0.5}$    & 26.0$_{_\pm1.0}$ \\ \cline{2-15} 
		&    & \checkmark &    &     &    & 27.8$_{_\pm1.0}$   & 34.4$_{_\pm0.8}$  & 41.3$_{_\pm1.0}$ & 19.6$_{_\pm0.8}$  & 31.9$_{_\pm1.2}$ & 12.2$_{_\pm2.1}$  & 18.3$_{_\pm1.2}$   & 29.2$_{_\pm0.5}$    & 26.9$_{_\pm0.5}$ \\ \cline{2-15} 
		&    &    & \checkmark &     &    & 27.4$_{_\pm0.8}$   & 36.3$_{_\pm1.1}$  & 39.7$_{_\pm0.9}$ & 19.4$_{_\pm0.9}$  & \textbf{34.8$_{_\pm1.5}$} & 10.0$_{_\pm2.0}$  & 19.6$_{_\pm1.1}$   & 30.3$_{_\pm0.7}$    & 27.2$_{_\pm0.3}$ \\ \cline{2-15} 
		& \checkmark & \checkmark &    &     &    & 27.8$_{_\pm0.4}$   & 36.4$_{_\pm0.4}$  & 39.4$_{_\pm1.0}$ & 18.1$_{_\pm0.2}$  & 33.8$_{_\pm1.5}$ & 10.9$_{_\pm1.9}$  & 18.8$_{_\pm1.3}$   & 30.1$_{_\pm0.2}$    & 26.9$_{_\pm0.6}$ \\ \cline{2-15} 
		& \checkmark & \checkmark & \checkmark &     &    & 25.7$_{_\pm0.9}$   & 36.3$_{_\pm1.1}$  & 40.4$_{_\pm0.7}$ & 20.1$_{_\pm0.3}$  & 34.5$_{_\pm1.3}$ & 12.8$_{_\pm2.2}$  & 24.1$_{_\pm1.6}$   & 30.3$_{_\pm0.4}$    & 28.0$_{_\pm0.5}$ \\ \cline{2-15} 
		&    &    &    & \checkmark  &    & 23.7$_{_\pm1.0}$   & 30.2$_{_\pm0.9}$  & 30.1$_{_\pm0.4}$ & 11.5$_{_\pm0.6}$  & 25.8$_{_\pm1.1}$ & 11.2$_{_\pm2.5}$   & 15.8$_{_\pm1.3}$   & 28.5$_{_\pm0.7}$    & 22.1$_{_\pm0.4}$ \\ \cline{2-15} 
		& \checkmark & \checkmark & \checkmark & \checkmark  &    & 26.7$_{_\pm0.6}$   & 36.2$_{_\pm1.2}$  & 41.0$_{_\pm0.6}$ & \textbf{20.3$_{_\pm0.7}$}  & 32.8$_{_\pm1.9}$ & \textbf{18.7$_{_\pm2.6}$}  & 21.1$_{_\pm1.4}$   & 29.8$_{_\pm0.6}$    & 28.3$_{_\pm0.5}$ \\ \cline{2-15} 
		&    &    &    &     & \checkmark & 23.5$_{_\pm0.7}$   & 29.0$_{_\pm0.6}$  & 27.1$_{_\pm0.5}$ & 10.9$_{_\pm0.2}$  & 23.2$_{_\pm1.0}$ & 9.8$_{_\pm2.6}$   & 16.0$_{_\pm1.4}$   & 26.4$_{_\pm0.2}$    & 20.8$_{_\pm0.8}$ \\ \cline{2-15} 
		& \checkmark & \checkmark & \checkmark & \checkmark  & \checkmark & \textbf{27.9$_{_\pm0.6}$}   & \textbf{37.8$_{_\pm0.6}$}  & \textbf{42.3$_{_\pm0.7}$} & 20.1$_{_\pm0.5}$  & 31.9$_{_\pm1.1}$ & 13.1$_{_\pm1.5}$  & \textbf{24.9$_{_\pm1.3}$}   & \textbf{30.6$_{_\pm0.9}$}    & \textbf{28.6$_{_\pm0.5}$} \\ \hline
		\textbf{UDA\_setting} \\
		~~ADDA~\cite{tzeng2017adversarial}                      &    &    &    &     &    & 25.7   & 35.8  & 38.5 & 12.6  & 25.2 & 9.1   & 21.5   & 30.8    & 24.9 \\
		~~DT+FT                     &    &    &    &     &    & 25.3   & 35.0  & 35.9 & 18.7  & 32.1 & 9.8   & 20.9   & \textbf{30.9}    & 26.1 \\
		~~FRCNN\_UDA~\cite{chen2018domain}                &    &    &    &     &    & 25.0   & 31.0  & 40.5 & \textbf{22.1}  & 35.3 & 20.2  & 20.0   & 27.1    & 27.6 \\ 
		~~Ours (SP only)                &    &    &    &     &    & \textbf{29.1}  & \textbf{39.7}  & \textbf{42.9} & 20.8  & \textbf{37.4} & \textbf{24.1}  & \textbf{26.5}   & 29.9    & \textbf{31.3} \\ \hline
	\end{tabular}
	\vspace{2pt}
	\caption{Quantitative results of our method on Scenario-5. 8 images(1 image per class) are sampled for each experiment round, and 1 object bounding box is annotated for corresponding class per image.}
	\label{CF_table}
	\vspace{-15pt}
\end{table*}

\vspace{-4mm}
\paragraph{Results for Scenario-1} As summarized in Table~\ref{SUC_table}, under FDA setting, comparing to source training model, the three different scaled image level adaptation modules independently provide favourable gain. Further combining them gives higher improvement (2.7 AP gain on mean value), indicating the complementary effect of alignments at different scales. The object instance level adaptation component independently generates 3.1 AP improvement. Combining image level components with instance level module further enhance the detector by 1.6 AP over instance level module only and 2.0 AP over the image level adaptation only, suggesting complementing effect of the two modules. Fune-tuning with the limited loosely annotated target samples brings minor improvement, but the gain is orthogonal to the adversarial adaptation modules. The combination of all proposed components brings 5.2 AP boost over the raw source model, which already outperforms state-of-art method~\cite{chen2018domain} under UDA setting.

It is clearly observable that baseline methods generate less improvement. The ADDA~\cite{tzeng2017adversarial} and FRCNN\_UDA~\cite{chen2018domain} methods barely brings any gains for the detector, suggesting they cannot effectively capture and mitigate the domain shift with only s few target data samples. The DT$ _{f} $+FT method results in about 1.0 AP gain, suggesting the style transfer method only weakly captures the domain shift in our setting where there is no such drastic style discrepancy as between those real images and comic or art works~\cite{johnson2016perceptual}. 

For the UDA setting, as sufficient target domain data are available, the three compared methods all get better results. While our proposed split pooling based adaptation brings much better results. We observe 6.4 AP gain over the baseline source model, indicating the module effectively captures and mitigates domain shift, for both cases where a few or sufficient target domain images are available.
\vspace{-5mm}
\paragraph{Result for other four scenarios} As presented in Table~\ref{SUC_table} to Table~\ref{CF_table}, for all the other scenarios, the results share similar trend with scenario-1. For FDA setting, our method provides effective adaptation for the source training model, significantly surpassing all baselines and outperforms state-of-the-art method under UDA setting. For UDA setting, our method generates SOTA performance with the proposed split pooling based adaptation.
It is interesting to note the performance of Scenario-1 (S$ \rightarrow  $U) is much lower than Scenario-3 (C$ \rightarrow  $U) though they share same test set. This is because the visual scene in SIM10K dataset is much simpler than that in Cityscapes, where more diverse car object instances are presented, providing better training statistics. Similar trend is observed in Scenario-2 and Scenario-4.

\begin{figure*}[!t]
	\centering
	\subfigure{ 
		\label{sc_unadapted_sample1} 
		\begin{minipage}{0.325\textwidth} 
			\centering 
			\includegraphics[width=1\linewidth, height=0.13\textheight]{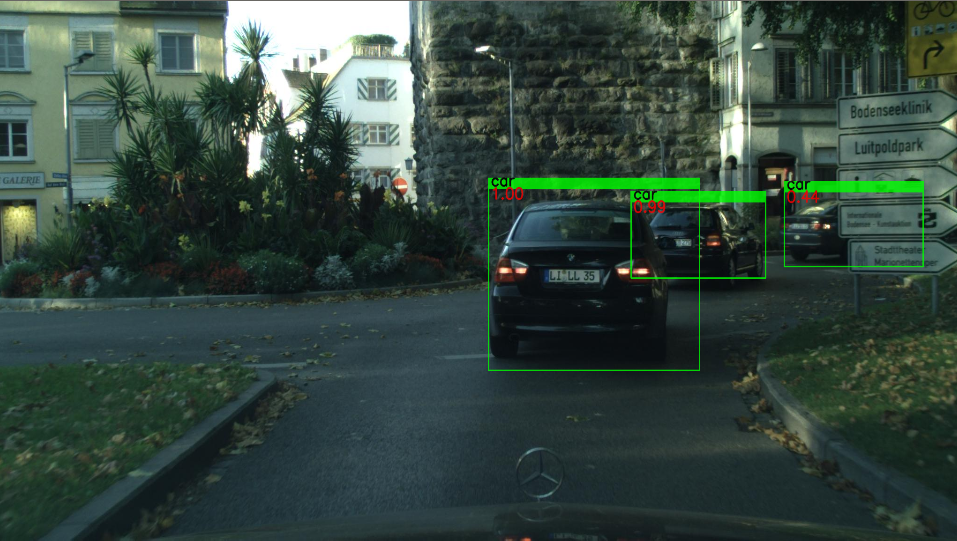} 
	\end{minipage}}%
	\subfigure{ 
		\label{sc_unadapted_sample2} 
		\begin{minipage}{0.325\textwidth} 
			\centering 
			\includegraphics[width=1\linewidth, height=0.13\textheight]{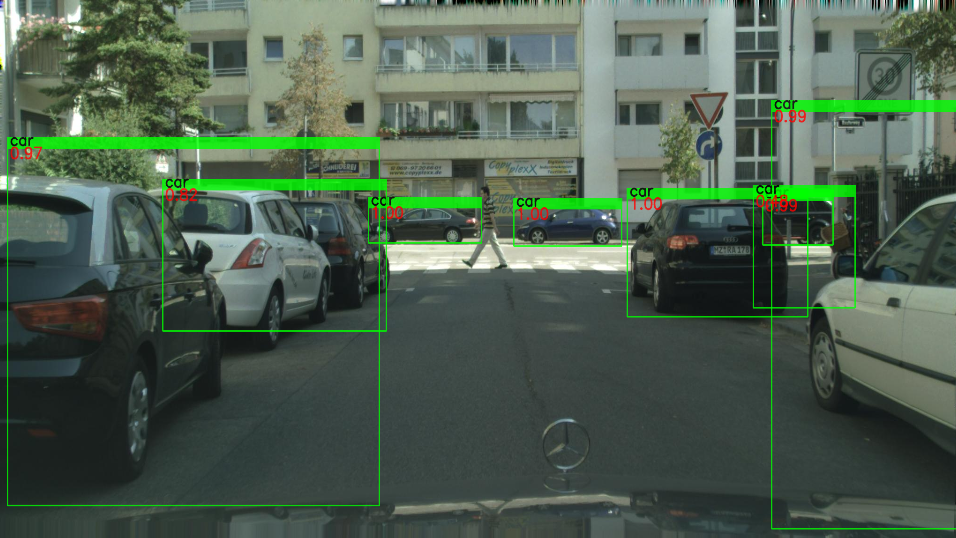}
	\end{minipage}}
	 \vspace{-8pt}
	 \hspace{-5pt}
	\subfigure{ 
		\label{sc_unadapted_sample3} 
		\begin{minipage}{0.325\textwidth} 
			\centering 
			\includegraphics[width=1\linewidth, height=0.13\textheight]{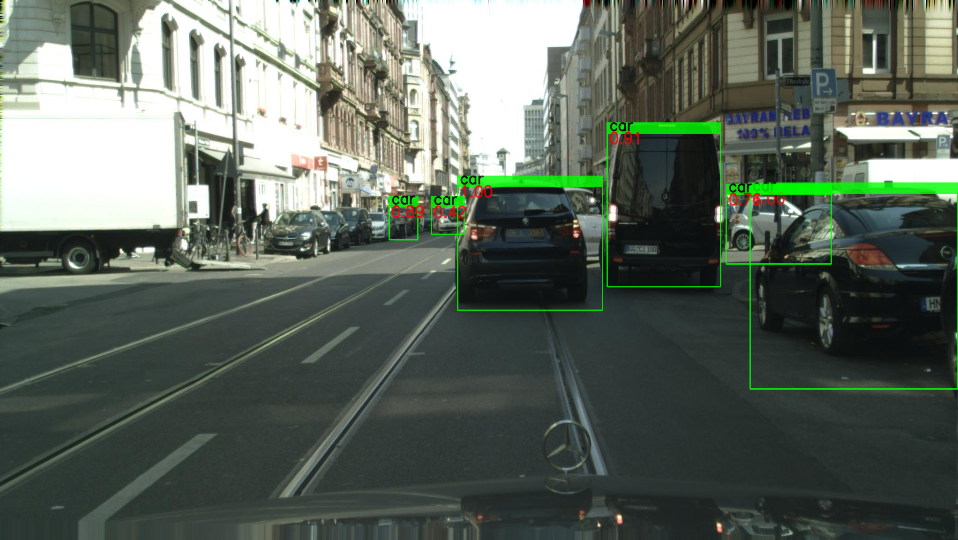} 
	\end{minipage}}
	\subfigure{ 
		\label{sc_adapted_sample1} 
		\begin{minipage}{0.325\textwidth} 
			\centering 
			\includegraphics[width=1\linewidth, height=0.13\textheight]{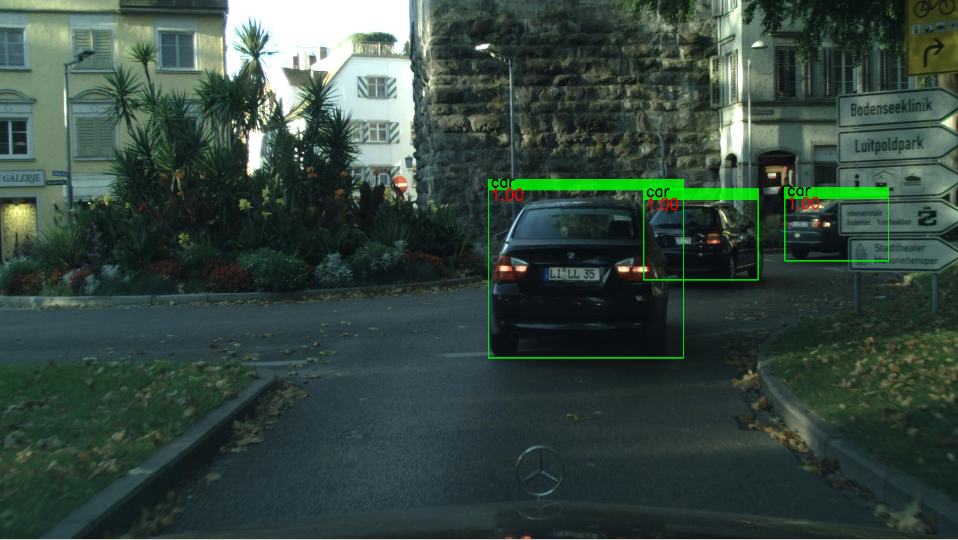} 
	\end{minipage}}%
	\subfigure{ 
		\label{sc_adapted_sample2} 
		\begin{minipage}{0.325\textwidth} 
			\centering 
			\includegraphics[width=1\linewidth, height=0.13\textheight]{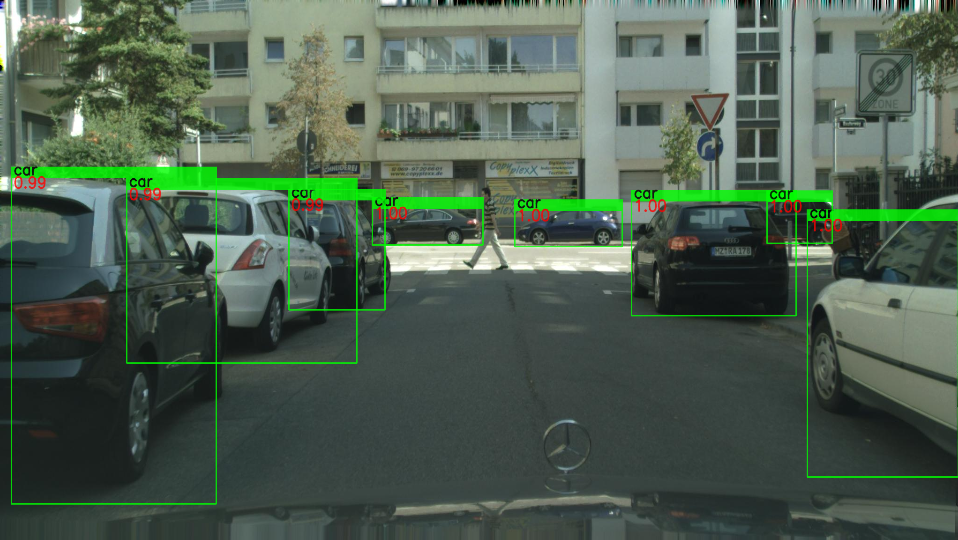} 
	\end{minipage}} 
	\hspace{-5pt}	
	\subfigure{ 
		\label{sc_adapted_sample3} 
		\begin{minipage}{0.325\textwidth} 
			\centering 
			\includegraphics[width=1\linewidth, height=0.13\textheight]{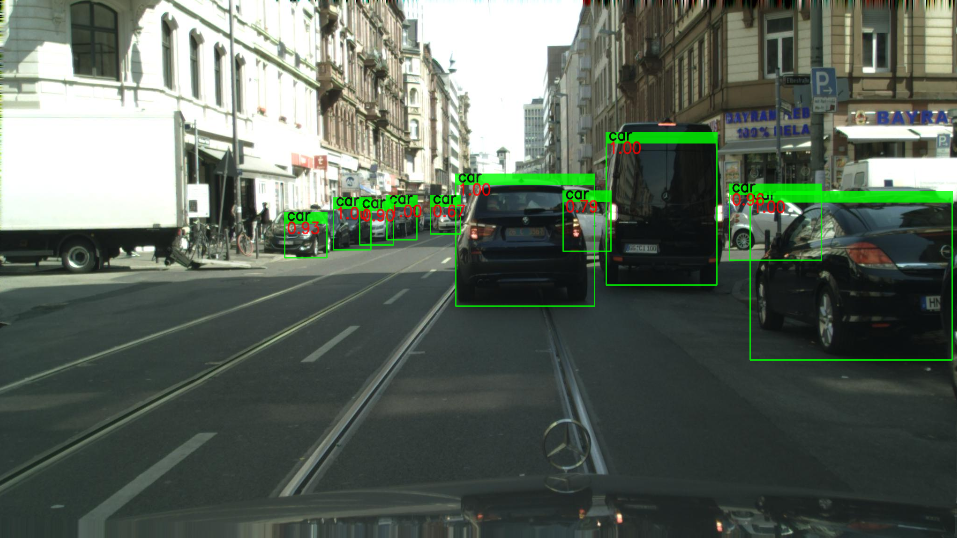} 
	\end{minipage}}
	
	\caption{Qualitative result. The results are sampled from S$  \rightarrow  $U scenario, we set a bounding box visualization threshold of 0.05. The first row are sample output from unadapted source training model, and second raw are corresponding detection output from adapted model.} 
	\label{qualitative_result} 
	\vspace{-15pt}
\end{figure*}

\begin{figure*}[!t]
\centering
\subfigure[]{ 
\label{vary_sample_number_SU} 
\begin{minipage}{0.28\textwidth} 
	\centering 
	\includegraphics[width=1\linewidth, height=0.135\textheight]{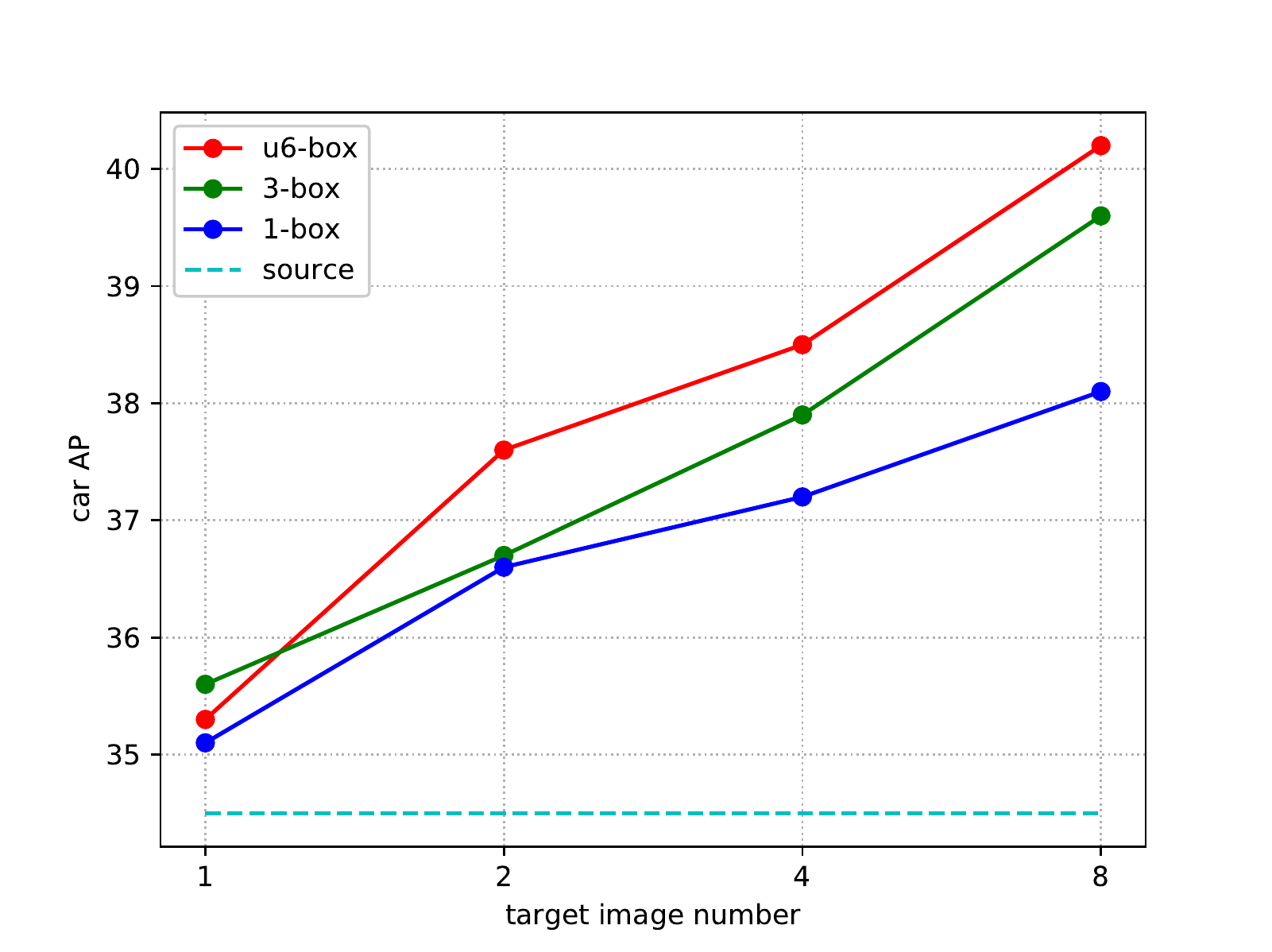} 
\end{minipage}}%
\subfigure[]{ 
	\label{vary_sample_number_UC} 
	\begin{minipage}{0.28\textwidth} 
		\centering 
		\includegraphics[width=1\linewidth, height=0.135\textheight]{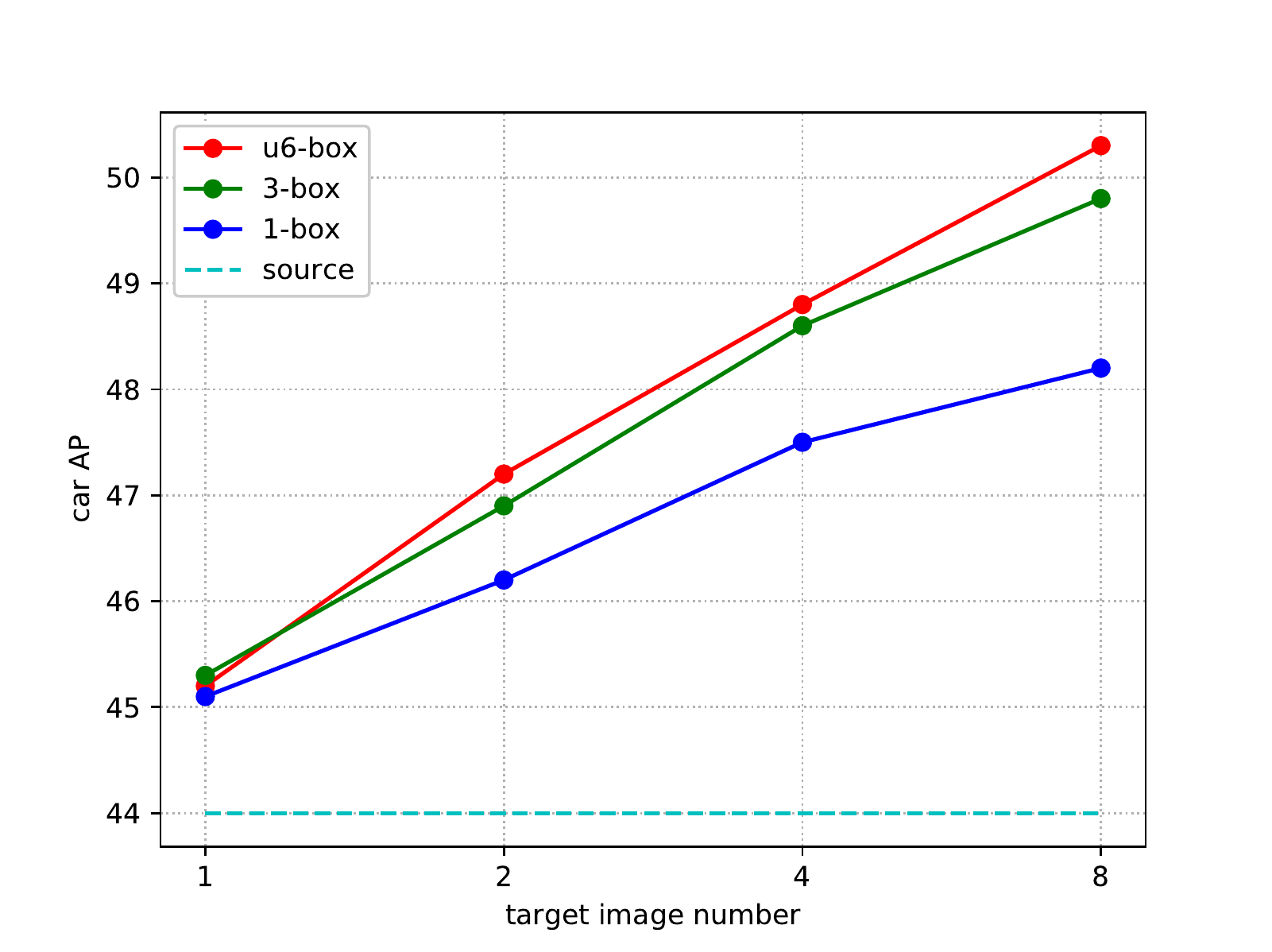} 
	\end{minipage}} 
	\subfigure[]{ 
		\label{vary_sample_number_CF} 
		\begin{minipage}{0.28\textwidth} 
			\centering 
			\includegraphics[width=1\linewidth, height=0.135\textheight]{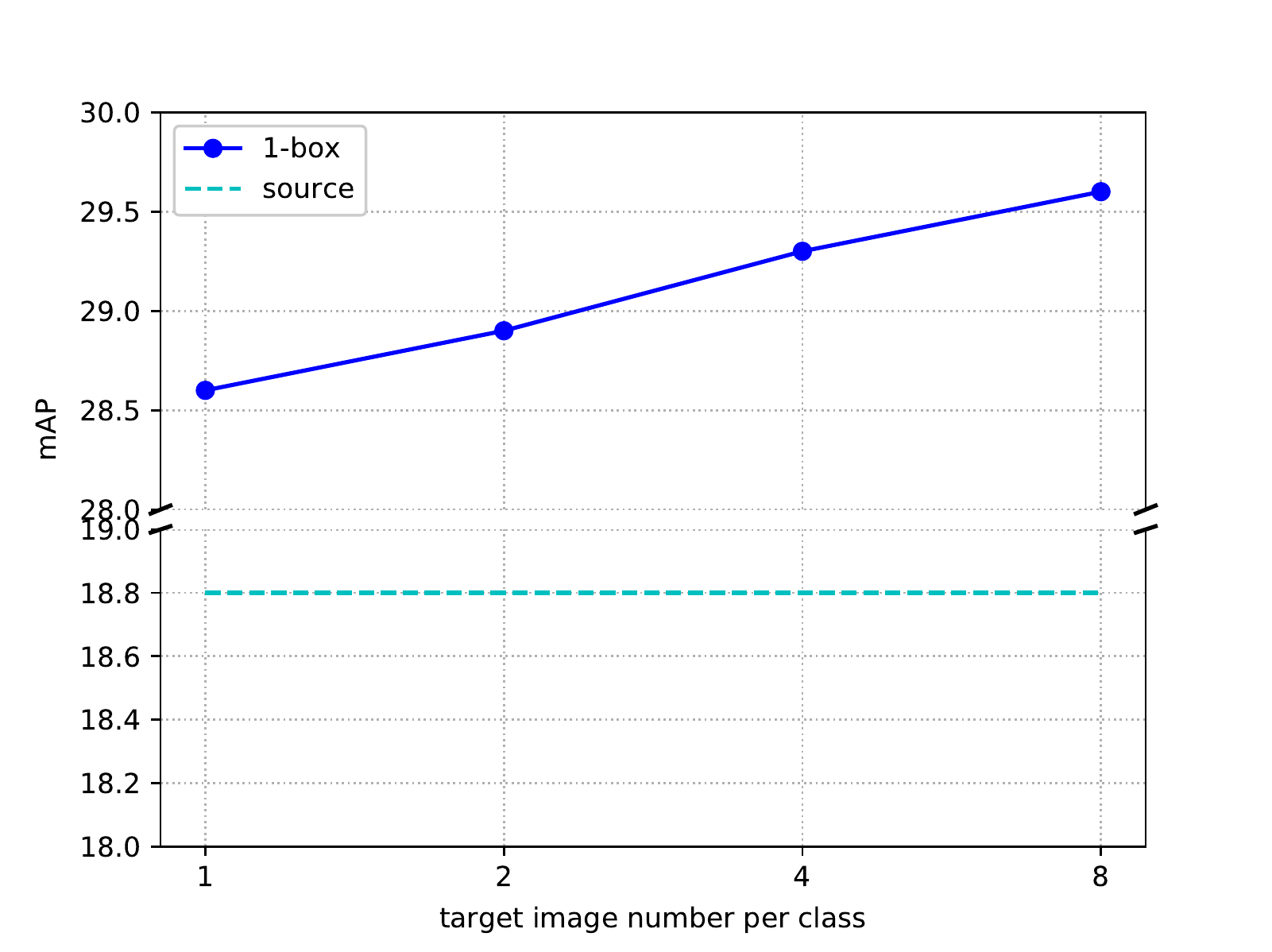} 
		\end{minipage}}
		\caption{Varying target sample image number and annotation boxes number. (a) S$ \rightarrow  $U. (b)  U$ \rightarrow  $C. (c)C$ \rightarrow  $F. 1-box, 3-box denote annotating only 1 or 3 box each sampled image, and u6-box means annotating at most 6 boxes as some images does not contain enough to 6 car objects.} 
		\label{vary_sample_number} 
		\vspace{-10pt}
	\end{figure*}


\subsection{Qualitative Results} \label{Qualitative_r}
Figure~\ref{qualitative_result} shows some qualitative result from Scenario 2 (S$ \rightarrow $C). It can be clearly observed that 1) the adapted model outputs tighter bounding boxes for each object, indicating better localization ability; 2) the adapted model places higher confidence on detected objects, especially for those harder objects (\emph{e.g.}, the car in the first image  occluded by the road sign); 3) the source model missed some small objects, while the adapted model can detect them.


\subsection{Ablation Analysis} \label{ablation}
\paragraph{Effect of pairing} As shown in Table~\ref{pairing_effect}, we independently examine the pairing effect on split pooling module and object instance level adaptation module. When not paired, we reduce input channel number of corresponding discriminator and remain the other parts unchanged. Without the introduced pairing, the performance of adaptation drops significantly. This indicates effectiveness of the pairing for augmenting the input data for discriminator learning.

\vspace{-4mm}

\paragraph{Number of sample images and annotated boxes} We examine the effect of varying the number of target domain images and annotating bounding boxes under Scenario-1, 4 and 5. We draw the mean value curve across all the sampling rounds. As car is abundant class for target domain of Scenario-1 and Scenario-4, we vary the annotated boxes number from 1 to up to 6 (at most 6 boxes considering a small set of images contain less than 6 car objects). We  vary the number of target   images   from 1  to 8 exponentially. For Scenario-5, as for most classes (like truck, bus, train, rider) there is only 1 instance in an image, we only annotate 1 box for each image. 
We do not examine beyond 8 images as there are already at most 48 (6 boxes* 8 images) and 64 (1 box*8 classes*8 images) object instances in Fig.4(a)(b) and Fig.4(c) involved, which can be deemed as sufficiently many for FDA evaluation.
As shown in Figure~\ref{vary_sample_number}, the results suggest common phenomenon that using more image and more box generates higher adaptation results. As image number increases exponentially, the roughly linear improvement suggests saturating effect.

\begin{table}[]
	\small
	\centering
	\renewcommand{\tabcolsep}{1.0pt}
	\renewcommand{\arraystretch}{1.1}
	\begin{tabular}{lcccccccc}
		\hline
		& sp$ _s $ & sp$ _m $ & sp$ _l $  & ins & S$  \rightarrow  $U & S$  \rightarrow  $C& C$  \rightarrow  $U & U$  \rightarrow C $ \\ \hline
		source     &&&&    & 34.1  & 33.5  & 44.5  & 44.0  \\
		pairing     & \checkmark      & \checkmark & \checkmark &     &  36.8$_{_\pm0.4}$  & 37.0$_{_\pm0.7}$  & 46.8$_{_\pm0.6}$  & 48.8$_{_\pm0.5}$ \\
		w/o  & \checkmark      & \checkmark & \checkmark &     &   34.8$_{_\pm0.5}$  & 34.3$_{_\pm0.6}$  & 44.6$_{_\pm0.3}$  & 45.8$_{_\pm0.4}$ \\
		pairing     &         &    &    & \checkmark  &   37.2$_{_\pm0.9}$  & 37.1$_{_\pm0.6}$  & 46.4$_{_\pm0.5}$  & 47.1$_{_\pm0.2}$ \\
		w/o  &  &    &    & \checkmark  &   35.7$_{_\pm0.6}$  & 34.9$_{_\pm0.5}$  & 44.1$_{_\pm0.6}$  & 45.3$_{_\pm0.8}$ \\
		pairing     & \checkmark      & \checkmark & \checkmark & \checkmark   &   39.3$_{_\pm0.6}$  & 39.8$_{_\pm0.7}$  & 48.4$_{_\pm0.7}$  & 50.6$_{_\pm0.5}$ \\
		w/o  & \checkmark      & \checkmark & \checkmark & \checkmark  &   36.1$_{_\pm0.6}$  & 36.8$_{_\pm0.6}$  & 44.5$_{_\pm0.3}$  & 45.5$_{_\pm0.4}$ \\ \hline
	\end{tabular}
	\vspace{2pt}
	\caption{The effect of the introduced pairing mechanism.}
	\label{pairing_effect}
	\vspace{-5pt}
\end{table}

\begin{table}[]
	\small
	\centering
	\renewcommand{\arraystretch}{1.1}
	\centering
	\begin{tabular}{llllllll}
		\hline
		& sp$ _{s} $ & sp$ _{m} $  & sp$ _{l} $  & ins & ft & mean & std \\ \hline
		source     &    &    &    &     &  & 33.5 & - \\ 
		SMFR     &    &    &    &     & \checkmark & \textbf{34.6} & \textbf{0.2} \\ 
		w/o &    &    &    &     & \checkmark & 30.1 & 1.8 \\ 
		SMFR     & \checkmark & \checkmark & \checkmark & \checkmark  &  & 39.6 & \textbf{0.3} \\ 
		w/o & \checkmark & \checkmark & \checkmark & \checkmark  &  & 39.4 & 2.1 \\ \hline
	\end{tabular}
	\vspace{5pt}
	\caption{The effect SMFR, with S$ \rightarrow $C scenario, mean and std denote mean and standard derivation of APs for the 10 runs.}
	\label{SMFR_effect}
\end{table}


\vspace{-4mm}

\paragraph{Sharing parameters among discriminators}
For split pooling based adaptation, we use the same discriminator architecture with shared parameters for different scale. While the discriminators could also be independent and not sharing parameters. As shown in Table~\ref{share_param_sp_pooling}, it is clearly observed that sharing the discriminator between small, medium and large scales provides much better results. Such interesting phenomenon suggests that image patches at different scales share similar representation characteristics for the image-level domain shift. They are complementary and combining them further strengthens the discriminator, resulting in better domain invariant representation.

\begin{table}[]
	\small
	\centering
	\renewcommand{\tabcolsep}{4.0pt}
	\renewcommand{\arraystretch}{1.1}
	\centering
	\begin{tabular}{lcccc}
		\hline
		& S$  \rightarrow  $U & S$  \rightarrow  $C& C$  \rightarrow  $U & U$  \rightarrow C $\\ \hline
		source         & 34.1  & 33.5  & 44.5  & 44.0  \\ 
		SP\_share      & 36.8$_{_\pm0.6}$  & 37.0$_{_\pm0.6}$  & 46.8$_{_\pm0.3}$  & 48.8$_{_\pm0.4}$  \\ 
		SP\_not\_share & 35.1$_{_\pm0.3}$  &  35.3$_{_\pm0.6}$ &  45.2$_{_\pm0.7}$ &  46.8$_{_\pm0.8}$ \\ \hline
	\end{tabular}
	\vspace{5pt}
	\caption{The effect of sharing/not sharing discriminator paramters between different scales' split pooling adaptation module.}	
	\label{share_param_sp_pooling}
	\vspace{-10pt}
\end{table}
\vspace{-3mm}
\paragraph{Stability gain from SMFR} Fine-tuning on small set of data unavoidably result in serve over-fitting, and instability is a common annoying feature of adversarial training. To evaluate the importance of the proposed source model feature regularization (SMFR), within one round of sample, we measure the standard derivation of the adapted model performance over 10 runs with different random parameter initialization. Table~\ref{SMFR_effect} illustrates that 1) Fine-tuning directly result in very large variance and suffer from severe overfitting, the tunned model performs worse than the source training model; Imposing SMFR drastically reduces variance, and the model actually benefits from the the limited target sample data. 2) While SMFR does not improve much of the overall performance of proposed components (\emph{i.e.}, sp$ _{s} $, sp$ _{m} $,sp$ _{l} $,ins), the variance is dramatically reduced.

\section{Conclusion}
In this paper, we explored the possibility of exploiting only few sample of target domain loosely annotated images to mitigate the performance drop of object detector caused by domain shift. Built on Faster R-CNN,  by carefully designing the adaptation modules and imposing proper regularization, our framework can robustly adapt a source trained model to target domain with very few target samples and still outperforms state-of-art methods accessing full unlabeled target set.
\vspace{-10pt}
\paragraph{Acknowledgement}							
Jiashi Feng was partially supported by NUS IDS R-263-000-C67-646,  ECRA R-263-000-C87-133 and MOE Tier-II R-263-000-D17-112.

{\small
\bibliographystyle{ieee_fullname}
\bibliography{egbib}
}

\end{document}